\pdfoutput=1
\documentclass[11pt]{article}
\usepackage[]{emnlp2022}
\usepackage{times}
\usepackage{latexsym}
\usepackage[T1]{fontenc}
\usepackage[utf8]{inputenc}
\usepackage{microtype}
\usepackage{inconsolata}
\usepackage{graphicx}
\usepackage{amsfonts,amssymb}
\usepackage[linesnumbered,ruled]{algorithm2e}
\usepackage{algpseudocode}
\usepackage{amsmath}
\usepackage{booktabs,chemformula}
\usepackage{multirow}
\usepackage{soul}
\usepackage{array}
\usepackage{xcolor}
\usepackage{url}

\newcommand{\red}[1]{\textcolor{red}{#1}}

\newcolumntype{P}[1]{>{\centering\arraybackslash}p{#1}}
\newcolumntype{M}[1]{>{\centering\arraybackslash}m{#1}}

\title{Generating Textual Adversaries with Minimal Perturbation}

\author{Xingyi Zhao \\
  Utah State University \\
  \texttt{xingyi.zhao@usu.edu} 
  \\\And
  Lu Zhang \\
  University of Arkansas\\
  \texttt{lz006@uark.edu}
  \\\AND
  Depeng Xu\\
  University of North Carolina at Charlotte\\
  \texttt{depeng.xu@uncc.edu}
  \\\And
  Shuhan Yuan\\
  Utah State University \\
  \texttt{shuhan.yuan@usu.edu}
}
  
\begin{document}
\maketitle

\begin{abstract}
Many word-level adversarial attack approaches for textual data have been proposed in recent studies. However, due to the massive search space consisting of combinations of candidate words, the existing approaches face the problem of preserving the semantics of texts when crafting adversarial counterparts. In this paper, we develop a novel attack strategy to find adversarial texts with high similarity to the original texts while introducing minimal perturbation. The rationale is that we expect the adversarial texts with small perturbation can better preserve the semantic meaning of original texts. Experiments show that, compared with state-of-the-art attack approaches, our approach achieves higher success rates and lower perturbation rates in four benchmark datasets\footnote{The code is available at \url{https://github.com/xingyizhao/TAMPERS}}.
\end{abstract}

\section{Introduction}
Recent studies have demonstrated that deep learning based natural language processing (NLP) models are vulnerable to adversarial attacks\cite{li2018textbugger,gao2018black, yang2020greedy}. Various textual adversarial  attack strategies have been proposed including character-level, word-level, and sentence-level attacks \cite{zhang2020adversarial,morris2020textattack}. In this paper, we focus on word-level attacks on text classifiers, especially the word substitution-based attacks, which are shown to be both effective and efficient \cite{alzantot2018generating,ren2019generating,zang2019word,jin2020bert,li2020bert,li2020contextualized}.

A successful adversarial attack should satisfy three requirements: 1) can fool the neural network models to make the wrong prediction; 2) the modification is slight and imperceptible to human judgments; and 3) can find the adversarial example in a reasonable run time. However, existing approaches suffer from various limitations. For example, the greedy algorithm-based attack approaches \cite{ren2019generating,jin2020bert} usually find a word in each step that can maximize the change of neural network outputs and do not explicitly consider the requirement of semantic preservation. On the other hand, the combinatorial optimization-based approaches, such as genetic algorithm \cite{alzantot2018generating,zang2019word}, define fitness functions to control the semantic quality of adversarial texts, but due to the huge search space, they usually need a large amount of time to find a solution, especially for a long text.

\begin{table}
\resizebox{\columnwidth}{!}{%
\begin{tabular}{|c|l|}
\hline
Original & \begin{tabular}[c]{@{}l@{}}A good film with a solid pedigree both in front of and,\\ more specifically, behind the camera.\end{tabular}\\ \hline
\begin{tabular}[c]{@{}c@{}}PWWS\\ (11.76\%)\end{tabular}       & \begin{tabular}[c]{@{}l@{}}A \textcolor{red}{\textbf{commodity}} film with a solid pedigree both in front of and,\\ more specifically, \textcolor{red}{\textbf{bum}} the camera.\end{tabular}
\\ \hline

\begin{tabular}[c]{@{}c@{}}TextFooler\\ (29.4\%)\end{tabular} & \begin{tabular}[c]{@{}l@{}}A \textcolor{red}{\textbf{better}} film with a \textcolor{red}{\textbf{stable}} pedigree both in \textcolor{red}{\textbf{newsweek}} of and,\\  more \textcolor{red}{\textbf{concretely}}, behind the \textcolor{red}{\textbf{camcorder}}.\end{tabular}                                    \\ \hline

\begin{tabular}[c]{@{}c@{}}Bert-Attack\\ (29.4\%)\end{tabular} & \begin{tabular}[c]{@{}l@{}}A \textcolor{red}{\textbf{fair}} \textcolor{red}{\textbf{positive}} with a solid pedigree both in \textcolor{red}{\textbf{after}} of and,\\ more \textcolor{red}{\textbf{roughly}}, behind the \textcolor{red}{\textbf{canvas}}.\end{tabular}                                   \\ \hline

\begin{tabular}[c]{@{}c@{}}TAMPERS\\ (5.88\%)\end{tabular}     & \begin{tabular}[c]{@{}l@{}}A \textcolor{red}{\textbf{harmless}} film with a solid pedigree both in front of and,\\ more specifically, behind the camera.\end{tabular}    \\ \hline

\end{tabular}
}
\caption{Adversarial examples generated by different attack approaches on MR. The bold red font indicates the perturbed words. The number in the bracket indicates the perturbation rate for each approach}
\label{tb:case}
\end{table}

In this work, we propose \textbf{T}extual  \textbf{A}dversarial attack with \textbf{M}inimal  \textbf{PE}rturbation in a \textbf{R}educed Search \textbf{S}pace (TAMPERS) to craft high quality adversarial texts.

The major motivation is that, by substituting a vulnerable word with its synonyms or sememes, we expect that the adversarial text with small perturbation can have large semantic preservation. We model the word-level attacks as a combinatorial optimization problem \cite{zang2019word}. We then develop a two-step heuristic for solving this problem. In the first step \textit{search space reduction}, we first build candidate lists for all the content words by combining synonyms from WordNet \cite{fellbaum2010wordnet} and sememes from HowNet \cite{dong2006hownet} similar to other word level attack strategies \cite{ren2019generating,zang2019word}. To avoid searching in a huge search space, instead of searching through all content words to generate adversarial texts, we reduce the search space by finding a small set of vulnerable words that are sufficient to fool the neural network using a greedy algorithm. Then, in the second step \textit{iterative search}, we further minimize the perturbation by iteratively restoring the substituted words back to the original words while keeping the classifier making the wrong prediction. Each time we restore a word, we adopt the genetic algorithm (GA) to search adversarial texts over the remaining substituted words. 
In this way, we expect to find an adversarial example with the least perturbation on the original text.

We have conducted experiments on four benchmark datasets.
The results show that our approach can successfully fool the fine-tuned BERT model with high success attack rates and low perturbation rates. Table \ref{tb:case} shows an example of adversarial texts generated by different attack approaches. We can notice that TAMPERS has a much lower perturbation rate compared with PWWS, TextFooler, and BERT-Attack. 

\section{Methodology}
Given any text $X=\{w_1,\ldots,w_n,\ldots,w_N\}$ with $N$ words and the true label $Y$, consider a pre-trained classifier that maps $X$ to $Y$, i.e., $F: X\rightarrow Y$. 
We assume the soft-label black-box setting where adversaries can query the classifier for classification probabilities on the given sample, but have no access to the model structures, parameters as well as training data. Our goal is to craft an adversarial sample $X^*$ such that $F(X^*)\neq Y$ with minimal word substitutions. 

\begin{figure}[t]
\centering
\includegraphics[scale=0.47]{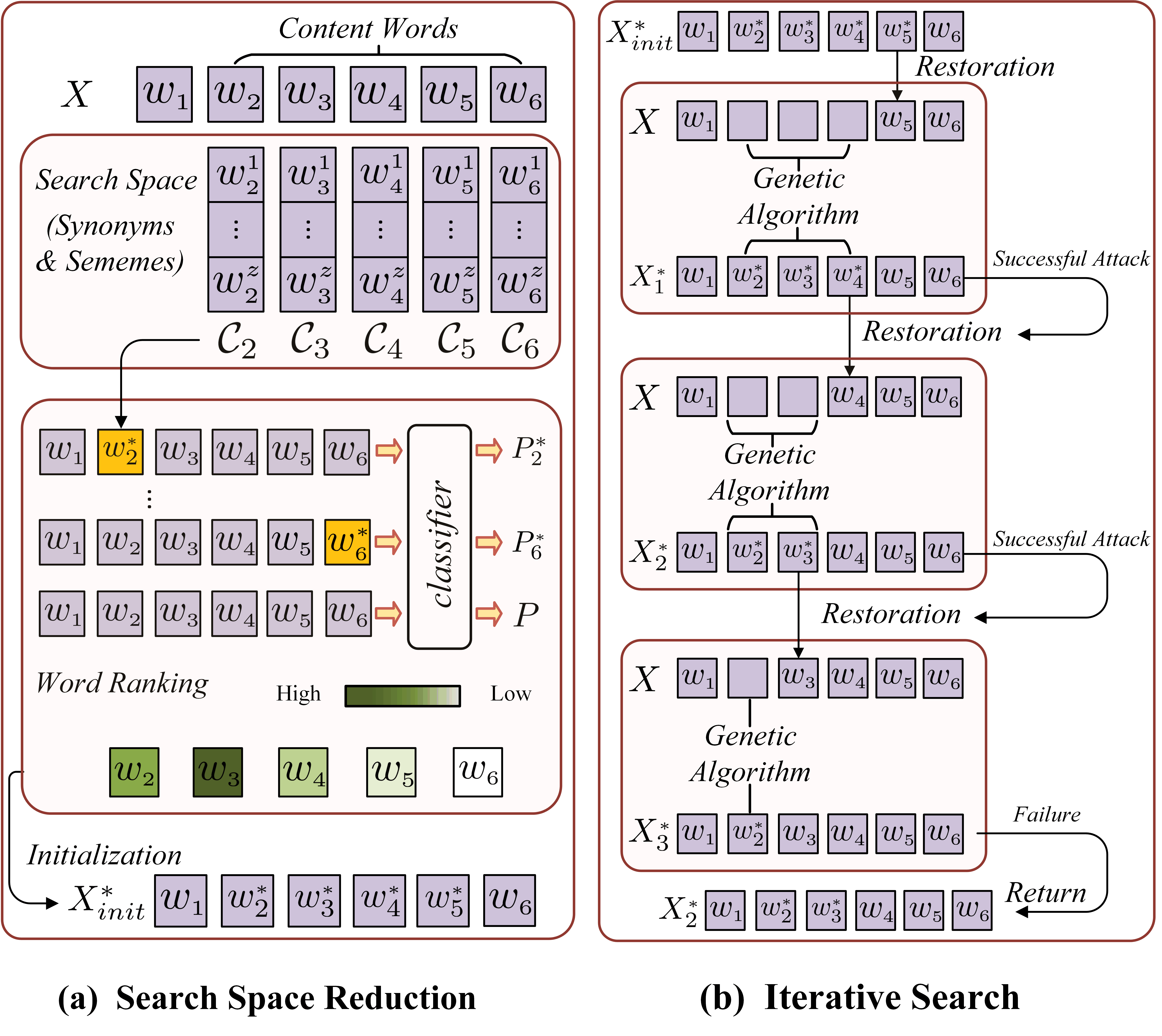}
\caption{Overview of TAMPERS}
\label{fig:framework}
\end{figure}

We propose a novel framework called \textbf{T}extual  \textbf{A}dversarial attack with  \textbf{M}inimal  \textbf{PE}rturbation in a \textbf{R}educed search \textbf{S}pace (TAMPERS), which consists of two steps.
The search space reduction step is to find vulnerable words with the goal of reducing the search space, while the iterative search step is to further minimize the perturbation so that the semantics of the original text can be well-preserved.
The overview of TAMPERS is shown in Figure \ref{fig:framework}. Below we describe each step in detail.

\subsection{Search Space Reduction}

In order to formalize the search space to craft the adversarial texts, the first step is to find the vulnerable words in the text $X=\{w_1,\ldots,w_n,\ldots,w_N\}$. Some related studies (\citealp{jin2020bert,li2020bert}) define the importance score of a content word $w_n$ based on the probability change of the classifier with and without the word $w_n$ (word deletion). Differently, we derive the importance score of the word $w_n$ by replacing the word with another word in a candidate list that leads to the maximum probability change of the classifier (word derivative).
The motivation of our approach is to approximate the computation of the gradient of the classifier output on the search space.

Specifically, given a content word $w_n$, we first find its synonyms from WordNet and sememes from HowNet with the same part of speech. Then, the substitution candidate list $\mathcal{C}_n$ for the word $w_n$ consists of the top-$z$ synonyms and sememes that are similar to $w_n$ based on the GloVe embeddings \cite{pennington2014glove}. After getting the candidate list $\mathcal{C}_n$, we use each word $w'_n \in \mathcal{C}_n$ to substitute $w_n$ to get a perturbed text $X'_n$. The set of perturbed texts formed by all words in the candidate list is denoted by $\mathcal{X}'_n$. Then, 
we compute the probability change of the classifier and define the importance score $S(w_n)$ as the maximum probability change across $\mathcal{X}'_n$:
\begin{equation}
    S(w_n)=\max\limits_{X'_n \in \mathcal{X}'_n} \{P(Y_{true}|X)-P(Y_{true}|X'_n)\}.
    \nonumber
\end{equation}
We denote the candidate word that achieves the score $S(w_n)$ as $w^*_n$.

Next, we rank all content words according to the importance score in descending order and adopt a greedy algorithm to generate the adversarial text, denoted as $X^*_{init}$. The greedy algorithm substitutes the content words one by one according to a descending order of their importance scores until the perturbed text can fool the classifier. In this step, we always substitute each word $w_n$ with the candidate word $w^*_n$ that is computed above.
After this step, we get a small set of vulnerable words $\mathcal{L}= \{w^*_{(1)},\ldots,w^*_{(k)},\ldots,w^*_{(K)}\}$ as the initial substitution words that is sufficient to fool the classifier, where $w^*_{(k)}$ is the $(k)$-th vulnerable word by importance score. The pseudo-code of the search space reduction is shown in Algorithm \ref{algr:ssr} in the appendix.

\subsection{Iterative Search}
Step 1 could significantly reduce the search space and produce an initial solution $X^*_{init}$ that already has a small perturbation. However, since the first step is in principle a one-step greedy strategy, we would like to further minimize the perturbation in order to better preserve the semantic integrity.
To this end, we develop a novel iterative search approach, where we iteratively restore the substituted words back to the original words according to the importance scores while keeping the classifier getting fooled. Each time after a word is restored, we reduce the list $\mathcal{L}$ accordingly and adopt GA to search for new substitutions for the perturbed words in the reduced $\mathcal{L}$. Since the search space of the GA is limited to $\mathcal{L}$, TAMPERS is much more efficient than previous GA-based approaches (e.g., \citealp{alzantot2018generating}). 

Symbolically, we start from the least vulnerable word $w^*_{(K)}$ in $\mathcal{L}$ and restore it to the original word $w_{(K)}$. Then, we run the GA over $\mathcal{L}\setminus w^*_{(K)}$ in order to find an adversarial text that can still fool the classifier. If we successfully find a solution, we then continue to restore the second least vulnerable word $w^*_{(K-1)}$ to $w_{(K-1)}$, reduce the list $\mathcal{L}$, and run the GA to find the adversarial text. We repeat this process until the GA cannot find a solution in the reduced search space. Finally, the algorithm returns the last successful solution as the output, which is the adversarial text with the minimal perturbation among those that have ever been found by the algorithm. The pseudo-code of the algorithm is shown in Algorithm \ref{algr:so} in the appendix.

The details of the genetic algorithm including initialization, selection, crossover, and mutation are described as follows

{\bf Step 1. Initialization.} This step is to generate the first generation of genetic algorithm $\mathcal{G}^0$ with $M$ samples (population size). Given the current set of vulnerable words $\mathcal{L}$, we randomly choose a candidate from $\mathcal{C}$ for each vulnerable word in $\mathcal{L}$ and generate $M$ texts, denoted as $\mathcal{G}^0 = \{X^0_1,\ldots,X^0_m,\ldots,X^0_M\}$. 

{\bf Step 2. Selection.} In the $g$-th generation $\mathcal{G}^g$, we first sort the texts based on the fitness score $h(X^g_m)$, defined as
\begin{equation}
        \resizebox{1\hsize}{!}{$
        h(X^g_m)=
        \begin{cases}
        P(Y_{true}|X)-P(Y_{true}|X^g_m),  &\mbox{if}\ F(X^g_m) = Y_{true}\\
        1,  &\mbox{if}\ F(X^g_m) \neq Y_{true}\\
        \end{cases}
        $}
        \label{eq:fitness}
    \nonumber
\end{equation}

Then, we select texts with the top 20\% highest fitness score in the current generation as elitism and directly send to the next generation $\mathcal{G}^{g+1}$.

{\bf Step 3. Crossover.}
For the texts in the $g$-th generation $\mathcal{G}^g$ with the top 50\% highest fitness score, we first randomly select two texts, denoted as $X^g_{p}$ and $X^g_{q}$. Then, for the positions of vulnerable words in the text, we randomly sample a word from either $X^g_{p}$ or $X^g_{q}$ at each position to generate a new text. We repeat the above two steps to generate the rest 80\% of texts in the next generation $\mathcal{G}^{g+1}$.

{\bf Step 4. Mutation.}
During the process of crossover, there is a small probability that we randomly select a new word from the candidate list instead of the word from $X^g_{p}$ or $X^g_{q}$.

After initializing the first generation $\mathcal{G}_0$ in Step 1 and repeating Steps 2 to 4 multiple iterations, if we are able to find an adversarial text giving current vulnerable words $\mathcal{L}$, it means compared with the adversarial text $X^*_{init}$ with a perturbation rate $K/N$, we generate new adversarial text with a perturbation rate $(K-1)/N$.

\section{Experiments}

\subsection{Experimental Setup}

{\bf\noindent Datasets and Victim Model.}
We evaluate TAMPERS on four text classification datasets, including \textbf{IMDB} \cite{maas-EtAl:2011:ACL-HLT2011}, \textbf{Yelp} \cite{zhang2015character}, \textbf{MR} \cite{pang2005seeing}, and \textbf{SST2} \cite{socher2013recursive}.
Following \citealp{jin2020bert} and \citealp{li2020bert}, we report the average outcome of 5 runs where in each run we randomly sample 1000 texts from each dataset. The victim model is the fine-tuned $\text{BERT}_{\text{base}}$.

\begin{table*}[t]
\small \centering
\begin{tabular}{|c|c|c|c|c|c|c|}
\hline
Dataset                                                                  & Model      & Original Acc.           & Attacked Acc.                   & Success Rate                     & Perturb Rate                   & Semantic Similarity                   \\ \hline
\multirow{4}{*}{\begin{tabular}[c]{@{}c@{}}IMDB\\ (225)\end{tabular}}    & TextFooler  & \multirow{4}{*}{94.7} & 0.8                           & 99.1                           & 8.7                            & 91.4                          \\ \cline{2-2} \cline{4-7} 
                                                                         & PWWS        &                         & 1.7                          & 98.2                            & 5.1                            & 87.9                           \\ \cline{2-2} \cline{4-7} 
                                                                         & Bert-Attack &                         & 3.3                          & 96.5                          & 6.6                            & 91.7                          \\ \cline{2-2} \cline{4-7} 
                                                                         & TAMPERS     &                         & \textbf{0.3} & \textbf{99.6} & \textbf{2.3}  & \textbf{93.5} \\ \hline
\multirow{4}{*}{\begin{tabular}[c]{@{}c@{}}Yelp\\ (129)\end{tabular}}    & TextFooler  & \multirow{4}{*}{97.3}   & 5.8                           & 94.0                           & 10.6                            & 88.5                           \\ \cline{2-2} \cline{4-7} 
                                                                         & PWWS        &                         & 5.5                          & 94.3                           & 7.5                            & 87.9                           \\ \cline{2-2} \cline{4-7} 
                                                                         & Bert-Attack &                         & 5.9                           & 93.9                           & 6.8                            & \textbf{89.7} \\ \cline{2-2} \cline{4-7} 
                                                                         & TAMPERS     &                         & \textbf{4.1} & \textbf{95.9} & \textbf{4.3}  & 89.1                           \\ \hline
\multirow{4}{*}{\begin{tabular}[c]{@{}c@{}}MR\\ (19.16)\end{tabular}}    & TextFooler  & \multirow{4}{*}{96.6} & 7.0                           & 92.8\                           & 18.7                           & 83.6                           \\ \cline{2-2} \cline{4-7} 
                                                                         & PWWS        &                         & 14.7                          & 84.8                           & 14.5                          & 84.9                         \\ \cline{2-2} \cline{4-7} 
                                                                         & Bert-Attack &                         & 11.3                          & 88.3                          & 14.8                           & 84.1                           \\ \cline{2-2} \cline{4-7} 
                                                                         & TAMPERS     &                         & \textbf{6.4}               & \textbf{93.4}                    &\textbf{9.6}                    & \textbf{86.3}                                     \\ \hline
\multirow{4}{*}{\begin{tabular}[c]{@{}c@{}}SST-2\\ (16.78)\end{tabular}} & TextFooler  & \multirow{4}{*}{93.8} & \textbf{5.2} & \textbf{94.5} & 17.2                           & 83.9                           \\ \cline{2-2} \cline{4-7} 
                                                                         & PWWS        &                         & 13.6                          & 85.6                           & 15.2                          & 84.0                          \\ \cline{2-2} \cline{4-7} 
                                                                         & Bert-Attack &                         & 7.0                           & 92.5                           & 14.0                          & \textbf{85.0}                          \\ \cline{2-2} \cline{4-7} 
                                                                         & TAMPERS     &                         & 6.3                           & 93.5                          & \textbf{10.1} & 84.9 \\ \hline
\end{tabular}
\caption{Results (shown in percentage) of attacking against fine-tuned BERT model on various datasets. We report the average outcome of 5 runs where in each run we randomly sample 1000 texts from each dataset. The number in the bracket indicates the average text length in each dataset.}
\label{tb:main}
\end{table*}

\begin{table}[]\small
\centering
\begin{tabular}{|c|c|c|}
\hline
Dataset               & Model   & Runtime (s/sample) \\ \hline
\multirow{3}{*}{IMDB} & Genetic Attack      & 2477 \\ \cline{2-3} 
                      & PSO     & 1495 \\ \cline{2-3} 
                      & TAMPERS & 98.4  \\ \hline
\end{tabular}
\caption{Runtime Comparison}
\label{tb:time}
\end{table}

{\bf\noindent Baselines.}
We choose three word-level adversarial attack models as baselines: 1) \textbf{TextFooler} adopts counter-fitting word embeddings and the greedy search to generate adversarial texts \cite{jin2020bert}; 2) \textbf{PWWS} is a synonym-based attack method using the greedy algorithm \cite{ren2019generating};  3) \textbf{BERT-Attack} uses a mask language model to predict candidate words and applies the greedy algorithm to craft adversarial texts \cite{li2020bert}. 

{\bf\noindent Evaluation Metrics.}
Following TextDefender (\citealp{li2021searching}), we adopt various evaluation metrics to evaluate the quality of adversarial texts, including \textit{original accuracy}, \textit{attacked accuracy}, \textit{attack success rate}, \textit{average perturbation rate}, and \textit{semantic similarity}. Following  TextFooler and BERT-Attack, we adopt Universal Sentence Encoder \cite{cer2018universal} to evaluate the semantic similarity between the original text and its adversarial counterpart.

{\bf\noindent Implementation Details.}
We constrain the size of candidates set $\mathcal{C}_n$ as 50 for each content word. For the initialized adversarial texts $X^*_{init}$ with more than one perturbed word, we apply iterative search to find better adversarial examples with fewer perturbation. In GA, we set the population size $M=10$ and maximal generation $T=100$.

\subsection{Experimental Results}

{\bf\noindent Attack Results.}
As shown in Table \ref{tb:main}, TAMPERS successfully attacks the fine-tune $\text{BERT}_{\text{base}}$ model on four datasets. In general, TAMPERS has the smallest perturbation rate and highest success attack rate among all baselines. As expected, the performance of TAMPERS is better on long texts. For the IMDB and Yelp datasets that mainly contain long texts, TAMPERS significantly reduces the perturbation rates compared with baselines and still achieves the highest attack success rate. On the MR and SST-2 datasets that mainly contain short texts, the improvement of TAMPERS is less significant but it can still reduce the perturbation rates by substituting around 2 words while maintaining high attack success rates.



{\bf \noindent Runtime Comparison.}
We compare the runtime of TAMPERS with that of Genetic Attack \cite{alzantot2018generating} and Particle Swarm Optimization (PSO) \cite{zang2019word}, both of which are also combinatorial optimization-based approaches. The results on the IMDB dataset are shown in Table \ref{tb:time}. We observe similar results on other datasets which are included in the appendix. As can be seen, TAMPERS is much faster and can craft an adversarial text in a reasonable time frame compared with Genetic Attack and PSO, thanks to the search space reduction step that significantly improves the efficiency of the algorithm.



\section{Conclusions}
In this work, we developed a novel word-level attack framework, called TAMPERS, that adopts a two-step approach with the goal of minimizing the perturbation. The first step reduces the search space and finds an initial adversarial text with the reduced perturbation. The second step then further minimizes the perturbation by iteratively restoring the substituted words back to the original words while keeping the classifier getting fooled. Experimental results show that TAMPERS generally achieves minimal perturbation in crafting adversarial examples while keeping a high success attack rate.



\section*{Limitations}
One limitation of the proposed framework is that in the iterative search step, each time we restore a vulnerable word $w^*_{(k)}$ to the original word $w_{(k)}$, GA starts from the scratch to find an adversarial example, which requires a lot of queries to the victim models. In the future, we would like to study how to improve query efficiency while maintaining the high success rate and low perturbation rate. 

\section*{Acknowledgement}
This work was supported in part by NSF 2103829.


\clearpage
\appendix

\section{Appendix}
\label{sec:appendix}

\subsection{Pseudo-code}
\label{sec:append:code}
The pseudo-code of Search Space Reduction and Iterative Search is given in Algorithms \ref{algr:ssr} and \ref{algr:so}, respectively.

\begin{algorithm}[h!]\small
            \SetKwInOut{Input}{Input}
            \SetKwInOut{Output}{Output}
            \caption{Search Space Reduction}
            \label{algr:ssr}
                \Input{Text $X$; Label $Y$; Classifier $F$} 
                \Output{Adversarial example $X^*_{init}$}
                    \For{each content word $w_n$ in $X$} { 
                      Get a substitution candidates set $\mathcal{C}_n$ for $w_n$\;
                     Compute the importance score $S(w_n)$ and get the candidate word $w^*_n$\;
                    }
                    Build a list $S$ of content words $w_n \in X$ in a descending order based on $S(w_n)$\;
                    
                    $\mathcal{L}=list()$\;
                    $X^* \gets \{w_1,...,w_n,...w_N\}$\;
                    \For{each word $w_n$ in $S$} {
                        $\mathcal{L}.append(w_n)$\; 
                        $X^*.replace(w_n,w^*_n)$\;
                        \If{$F(X^*) \neq Y$} {
                            $X^*_{init} \gets X^*$\;
                            \Return $X^*_{init}$ \bf{and} $\mathcal{L}$\;
                            }
                    }
                    \Return \textbf{None}\;
        \end{algorithm}

\begin{algorithm}[h!]\small
\SetKwInOut{Input}{Input}
\SetKwInOut{Output}{Output}
\caption{Iterative Search}
\label{algr:so}
        \Input{Text $X$; Label $Y$;  Classifier $F$; Adversarial text $X^*_{init}$; Vulnerable word list $\mathcal{L}$ with size $K$; Substitution candidate sets $Q$ for $w \in \mathcal{L}$}
        \Output{Adversarial example $X^*$}
            $X^* \gets X^*_{init}$\;
            \For{$k=K\ \text{to}\ 1$} {
                $\mathcal{L} = \mathcal{L}\setminus w_{(k)}$\;
                Initialize the first generation $\mathcal{G}^0$\;
                \For{$g=1\ \text{to}\ T$} {
                    Generate texts in $\mathcal{G}^{g}$ based on Selection, Crossover and Mutation\;
                    $\mathcal{G}^{g} \gets sort(\mathcal{G}^{g})$ \; 
                    \If{$F(\mathcal{G}^{g}[0]) \neq Y$} {
                        $X' \gets \mathcal{G}^{g}[0]$\;
                        \textbf{break}\;
                    } 
                }
                \If{$F(\mathcal{G}^{g}[0]) == Y$} {
                    \textbf{break}\;
                }       
                $X^* \gets X'$\;
            }
            \Return $X^*$\;
\end{algorithm}

\subsection{Implementation Details}
For TextFooler and PWWS, we follow the implementations in TextAttack (\citealp{morris2020textattack}), while for BERT-Attack, we follow the implementations in TextDefender (\citealp{li2021searching}). We use the fine-tuned victim models for all the datasets shared by TextAttack \footnote{\url{https://huggingface.co/textattack}}. All experiments are run on AMD Ryzen Threadripper 3960X 24-core Processor and NVIDIA GeForce RTX 3090. 

\begin{table*}[ht!]
\centering
\resizebox{\textwidth}{!}{  
\begin{tabular}{|c|c|c|c|c|c|c|c|}
\hline
Dataset                                            & Method  & Original Acc.                                & Attacked Acc. & Success Rate & Perturb Rate & Semantic Similarity & \begin{tabular}[c]{@{}c@{}}Runtime\\ (s/example)\end{tabular}   \\ \hline
\multirow{3}{*}{\begin{tabular}[c]{@{}c@{}}MR\\ (19.25)\end{tabular}}    & Genetic Attack                           & \multicolumn{1}{c|}{\multirow{3}{*}{97.2}} & 30.8                             & 68.3                           & 14.5                           & \textbf{86.5}                     & 133.9 \\ \cline{2-2} \cline{4-8} 
                                                                         & PSO                          & \multicolumn{1}{c|}{}                        & \textbf{6.0}                              & \textbf{93.8}                           & 20.3                           & 81.8                    & 25.2  \\ \cline{2-2} \cline{4-8} 
                                                                         & \multicolumn{1}{c|}{TAMPERS} & \multicolumn{1}{c|}{}                        & 10.0                              & 89.7                           & \textbf{8.3}                            & 85.9                   & \textbf{10.6}  \\ \hline \hline
\multirow{3}{*}{\begin{tabular}[c]{@{}c@{}}SST-2\\ (16.77)\end{tabular}} & Genetic Attack                           & \multirow{3}{*}{92.4}                      & 35.6                             & 61.5                           & 15.6                           & \textbf{86.3}                    & 43.2  \\ \cline{2-2} \cline{4-8} 
                                                                         & PSO                          &                                              & 8.4                              & 90.9                          & 19.8                           & 82.2                     & 14.4  \\ \cline{2-2} \cline{4-8} 
                                                                         & \multicolumn{1}{c|}{TAMPERS} &                                              & \textbf{6.8}                              & \textbf{92.7}                           & \textbf{10.4}                           & 84.9                    & \textbf{9.8}   \\ \hline
\end{tabular}
}
\caption{Attacking results (shown in percentage) of Genetic Attack, PSO, and TAMPERS against fine-tuned BERT.}
\label{tb:ga-pso}
\end{table*}

\subsection{Experimental Results}
Due to the high time-consuming of Genetic Attack (\citealp{alzantot2018generating}) and PSO (\citealp{zang2019word}) algorithms, especially on long texts, we compare Genetic Attack and PSO with TAMPERS on MR and SST-2 by randomly selecting 250 correctly predicted texts. Table \ref{tb:ga-pso} shows the experimental results. Overall, compared with Genetic Attack and PSO, TAMPERS can still achieve the lowest perturbation rate with a high success attack rate. Meanwhile, TAMPERS is much faster than Genetic Attack and PSO even in datasets with short texts, which shows the advantage of the search space reduction step proposed in our framework. 

\subsection{Case Study}
We show the cases of adversarial examples generated by different approaches on IMDB and Yelp in Tables \ref{tb:imdb} and \ref{tb:yelp}, respectively.

\begin{table*}[]
\small
\centering
\begin{tabular}{|P{0.1\linewidth}|P{0.80\linewidth}|}
\hline
\begin{tabular}[c]{@{}c@{}} IMDB\\ (Positive)\end{tabular}       &\begin{tabular}[c]{p{0.98\linewidth}}  A wonderful little production. The filming technique is very unassuming very old time BBC fashion and gives a comforting, and sometimes discomforting, sense of realism to the entire piece. The actors are extremely well chosen Michael Sheen not only ``has got all the polari'' but he has all the voices down pat too! You can truly see the seamless editing guided by the references to Williams' diary entries, not only is it well worth the watching but it is a terrificly written and performed piece. A masterful production about one of the great master's of comedy and his life. The realism really comes home with the little things: the fantasy of the guard which, rather than use the traditional 'dream' techniques remains solid then disappears. It plays on our knowledge and our senses, particularly with the scenes concerning Orton and Halliwell and the sets (particularly of their flat with Halliwell's murals decorating every surface) are terribly well done.  \end{tabular}                                                                                             \\ \hline
\begin{tabular}[c]{@{}c@{}}PWWS\\ (negative)\end{tabular}       & \begin{tabular}[c]{p{0.98\linewidth}}\red{ \bf amp} wonderful \red{ \bf petty} production. The filming technique is very unassuming very old time BBC fashion and gives a comforting, and sometimes discomforting, sense of realism to the entire \red{ \bf pick}. The actors are extremely well chosen Michael Sheen not only ``has got all the polari'' but he has all the voices down \red{ \bf slick} too! You can truly see the seamless editing \red{ \bf steer} by the references to Williams' diary entries, not only is it \red{ \bf substantially} worth the watching but it is a terrificly written and performed piece. A masterful production about \red{ \bf unitary} of the great master's of \red{ \bf clowning} and his \red{ \bf spirit}. The \red{ \bf pragmatism} really \red{ \bf amount} home with the little things: the \red{ \bf illusion} of the guard which, rather than use the traditional `dream' techniques remains solid then disappears. It plays on our knowledge and our senses, particularly with the \red{ \bf conniption} concerning Orton and Halliwell and the sets (particularly of their flat with Halliwell's murals decorating every surface) are \red{ \bf atrociously} \red{ \bf fountainhead} \red{ \bf through}.\end{tabular}                                                        \\ \hline
\begin{tabular}[c]{@{}c@{}}TextFooler\\ (negative)\end{tabular} & \begin{tabular}[c]{p{0.98\linewidth}}A \red{ \bf unbelievable} little production. The filming \red{ \bf tech} is \red{ \bf terribly} unassuming \red{ \bf incredibly} oldtimeBBC \red{ \bf attire} and \red{ \bf begs} a comforting, and \red{ \bf invariably} discomforting, \red{ \bf meanings} of \red{ \bf reality} to the \red{ \bf total} \red{ \bf lump}. The actors are \red{ \bf horribly} well taking Michael Sheen not only ``has got all the polari'' but he has all the voices down pat too! You can truly \red{ \bf suppose} the seamless editing \red{ \bf directorate} by the \red{ \bf referencing} to Williams’ diary \red{ \bf inscriptions}, not only is it \red{ \bf sufficiently valuable} the watching but it is a terrificly \red{ \bf  writing} and \red{ \bf achieve} piece. A masterful production about one of the \red{ \bf excellent} master’s of \red{ \bf  entertaining} and his \red{ \bf duration}. \red{ \bf Both} realism \red{ \bf surely} \red{ \bf arrives} \red{ \bf interiors} with the \red{ \bf scarcely} \red{ \bf stuff}: the \red{ \bf imagined} of the \red{ \bf guarding} which, \red{ \bf  reasonably} than \red{ \bf employing} the \red{ \bf routine} ‘dream’ \red{ \bf technician} remains solid then disappears. \red{ \bf He gaming} on our \red{ \bf proficiency} and our \red{ \bf wits}, particularly with the \red{ \bf imagery} \red{ \bf implicated} \red{ \bf Smackdown} and Halliwell and the \red{ \bf provides} (\red{ \bf substantially} of their flat with Halliwell’s \red{ \bf graphics} decorating every \red{ \bf  cosmetic}) are \red{ \bf  freakishly} \red{ \bf successfully }\red{ \bf effected}.\end{tabular} \\ \hline
\begin{tabular}[c]{@{}c@{}}BertAttack\\ (negative)\end{tabular} & \begin{tabular}[c]{p{0.98\linewidth}}A wonderful little production. The filming \red{ \bf work} is very unassuming very oldtimeBBC \red{ \bf sort} and gives a comforting, and \red{ \bf perhaps} discomforting, sense of \red{ \bf horror} to the entire \red{ \bf exhibition}. The actors are \red{ \bf seriously} \red{ \bf holy} \red{ \bf appointed} Michael Sheen not only ``has got all the polari'' but he has all the voices down pat too! You can truly see the seamless editing guided by the references to Williams' \red{ \bf newspaper} entries, not only is it well \red{ \bf enjoyed }the watching but it is a terrificly written and performed piece. \red{ \bf good} masterful production about \red{ \bf being} of the great master’s of \red{ \bf comedians} and his \red{ \bf personality}. The realism really \red{ \bf starts }home with the little things: the \red{ \bf projection} of the guard which, rather than use the traditional ‘dream’ techniques remains solid then \red{ \bf dismiss}. It plays on our knowledge and our senses, particularly with the \red{ \bf elements surrounding} Orton and Halliwell and the \red{ \bf production} (particularly of their flat with Halliwell’s \red{ \bf pupils }decorating \red{ \bf totally} surface) are terribly \red{ \bf thoroughly awful}.\end{tabular}                                                           \\ \hline
\begin{tabular}[c]{@{}c@{}}TAMPERS\\ (negative)\end{tabular}    & \begin{tabular}[c]{p{0.98\linewidth}}A wonderful \red{ \bf less} production. The filming technique is very unassuming very oldtimeBBC fashion and gives a comforting, and sometimes discomforting, sense of realism to the entire piece. The actors are extremely well chosen Michael Sheen not only "has got all the polari" but he has all the voices down pat too! You can truly see the seamless editing guided by the references to Williams’ diary entries, not only is it well worth the watching but it is a terrificly written and performed scrap . A \red{ \bf tyrannical preparation} about one of the great master’s of \red{ \bf drollery} and his life. The realism really comes home with the little things: the fantasy of the guard which, rather than use the traditional ‘dream’ techniques remains solid then disappears. It plays on our knowledge and our senses, particularly with the scenes concerning Orton and Halliwell and the sets (particularly of their flat with Halliwell’s murals decorating every surface) are \red{ \bf atrociously so} done.\end{tabular}                                                                                                   \\ \hline
\end{tabular}
\caption{Given an IMDB review, the adversarial examples generated by different approaches. The perturbed words are highlighted in red.}
\label{tb:imdb}
\end{table*}

\begin{table*}[]
\small
\centering
\begin{tabular}{|P{0.1\linewidth}|P{0.80\linewidth}|}
\hline
\begin{tabular}[c]{@{}c@{}}Yelp\\ (Negative)\end{tabular}       & \begin{tabular}[c]{p{0.98\linewidth}}I don't understand what all the hoopla is about. The food was lousy. The ribs tasted like the cow had mad cow disease. The sauce was weak. Overall, the food is overpriced. I've gone to this place twice and all I can say is I loved the beer!\end{tabular}           \\ \hline
\begin{tabular}[c]{@{}c@{}}PWWS\\ (Positive)\end{tabular}       & \begin{tabular}[c]{p{0.98\linewidth}}\red{\bf iodin} don't \red{\bf realise} what all the hoopla is about. The food was lousy. The ribs tasted like the cow had mad cow disease. The sauce was weak. Overall, the food is overpriced. I've \red{\bf extend} to this place twice and all I can say is \red{\bf iodin} loved the beer!\end{tabular}    \\ \hline
\begin{tabular}[c]{@{}c@{}}TextFooler\\ (Positive)\end{tabular} & \begin{tabular}[c]{p{0.98\linewidth}}I don't understand what all the \red{\bf tizzy} is about. The food was \red{\bf sorrowful}. The ribs tasted like the cow had mad cow disease. The sauce was weak. \red{\bf Comprehensive}, the food is overpriced. I've gone to this place twice and all I can say is I loved the \red{\bf brews}!\end{tabular} \\ \hline
\begin{tabular}[c]{@{}c@{}}BertAttack\\ (Positive)\end{tabular} & \begin{tabular}[c]{p{0.98\linewidth}}I don't \red{ \bf forget} what all the hoopla is about. The \red{ \bf there} was lousy. The ribs tasted like the cow had mad cow disease. The sauce was weak. Overall, the food is overpriced. I've gone to this place twice and all I can say is \red{\bf totally} loved the beer!                                                                                                                                                       \end{tabular}  \\ \hline
\begin{tabular}[c]{@{}c@{}}TAMPERS\\ (Positive)\end{tabular}    & \begin{tabular}[c]{p{0.98\linewidth}}I don't understand what all the hoopla is about. The \red{\bf diet} was lousy. The ribs tasted like the cow had mad cow disease. The sauce was weak. Overall, the \red{\bf snacks} is overpriced. I've \red{\bf coming} to this place twice and all I can say is I loved the beer!\end{tabular}       \\ \hline
\end{tabular}
\caption{Given a Yelp review, the adversarial examples generated by different approaches. The perturbed words are highlighted in red.}
\label{tb:yelp}
\end{table*}

\end{document}